\documentclass{article}

\usepackage[preprint,nonatbib]{neurips_2022}

\usepackage[utf8]{inputenc} 
\usepackage[T1]{fontenc}    
\usepackage{hyperref}       
\usepackage{url}            
\usepackage{booktabs}       
\usepackage{amsfonts}       
\usepackage{nicefrac}       
\usepackage{microtype}      
\usepackage{xcolor}         
\usepackage{graphicx}       

\newcommand{\envabbrev}{}

\title{\envabbrev A Cooperative Reinforcement Learning Environment for Detecting and Penalizing Betrayal}

\author{%
  Nikiforos Pittaras \\
  University of Athens \\
  \texttt{npittaras@di.uoa.gr}\\
}

\begin{document}

\maketitle

\begin{abstract}

  In this paper we present a Reinforcement Learning environment that leverages agent cooperation and communication, aimed at detection, learning and ultimately penalizing betrayal patterns that emerge in the behavior of self-interested agents. We provide a description of game rules, along with interesting cases of betrayal and trade-offs that arise. Preliminary experimental investigations illustrate a) betrayal emergence, b) deceptive agents outperforming honest baselines and b) betrayal detection based on classification of behavioral features, which surpasses probabilistic detection baselines.
  Finally, we propose approaches for penalizing betrayal, list directions for future work and suggest interesting extensions of the environment towards capturing and exploring increasingly complex patterns of social interactions.

\end{abstract}

\section{Introduction}

Establishing truthfulness in AI is a critical open problem in Safety and Alignment efforts \cite{evans2021truthful}. A powerful AI system that adopts strategies of deception and betrayal, i.e. manipulation of beliefs and prior assumptions of humans, may be a quick one-way ticket to a treacherous turn.
Detection and diagnosis of betrayal patterns is challenging;
poor explainability of black-box agents make it difficult to deduce intent, goals and beliefs by inspecting internal model workings and/or operational outputs \cite{samek2019towards,yampolskiy2020unexplainability}.
To make matters worse, intelligent agents capable of long-term strategizing would render human interpretation and recognition of suspicious patterns in action sequences very difficult. At the same time, instrumentally convergent attributes such as self-preservation and resistance to corrigibility could result in AI systems that deliberately utilize
obfuscation or exhibit deceptive alignment \cite{amodei2016concrete}, placing further obstacles in understanding their objectives.

In these settings, Anomaly Detection countermeasures \cite{pang2021deep} aim to identify, prevent, correct or mitigate adverse outcomes prior to system deployment.
For instance, betrayal detection and quantification can serve as tripwires and honeypots to avoid future harms, catching systems that exhibit problematic behavior early on \cite{amodei2016concrete}.
Additionally, betrayal penalization approaches aim to regularize agents away from undesirable actions during training. Ideally, this resolution should be interpretable to human evaluators and generalize well to different problems, agent architectures and domains, having efficiently internalized concepts of betrayal and deception.

Reinforcement Learning (RL) can provide a tractable avenue for investigating such scenarios \cite{dafoe2020open}, using environments where reliable reward accumulation heavily depends upon cooperation between agents and complex social interactions occur \cite{oroojlooyjadid2019review,chelarescu2021deception}. In this work, we adopt such an approach, focused on detecting and penalizing undesirable behaviors of deception and betrayal in a custom, communication-based navigation task.

\section{Related Work}

Previous studies have explored agent communication in a multiagent RL setting; Kajic et al. \cite{kajic2020learning} investigate message-based navigation similar to the proposed work, while Cao et al. \cite{cao2018emergent} study communication grounding with respect to game rules in agents of varying degrees of self-interest. In the work of Kim et al. \cite{kim2020communication}, agents used a world model to predict future agent intents and environment dynamics to generate, compress and transmit imagined trajectories.
Other works explore topological configurations different from fully-connected communication, such as the learnable hierarchical approach in Sheng et. al \cite{sheng2022learning}, while communication via noisy channels has been investigated in Tung et. al \cite{tung2021effective}.

Agent deception, betrayal, truthfulness and trustworthiness has been previously investigated in multiple settings \cite{chelarescu2021deception}; for instance, Christiano et al. \cite{elk} present a challenge of discovering latent knowledge in an agent that may produces false / unreliable reports, while Usui et al. \cite{usui2021symmetric} evaluate analytic solutions of different strategies in iterated Prisoner's Dilemmas.

Social dilemmas that gauge cooperation versus self-interest are explored in Leibo et al. \cite{leibo2017multi}, applied via games like ``Gather'' and ``Wolfpack''. ``Hidden Agenda'' is a team-based game offering a complex action set including 2D navigation, agent / environment interaction, deception and trustworthiness estimation via voting, and is investigated by Kopparapu et al. \cite{kopparapu2022hidden}. Asgharnia \cite{asgharnia2022learning} use a hierarchical fuzzy, situation-aware learning scheme to learn and utilize deception against one or multiple adversaries in a custom environment.

Mitigation approaches include the work in Hughes et al. \cite{hughes2018inequity}, where reward regularization is approached by adding an inequity penalty in games with short-term versus long-term dilemmas, like ``Cleanup'' and ``Harvest''. Jaques et al.  \cite{jaques2019social} use the same setting with a mutual information-based mechanism that favors influential communication between agents, adopting a correlation assumption of influence to cooperation.
Blumenkamp et al. \cite{blumenkamp2020emergence} utilize cooperative policy learning via shared differentiable communication channel in three custom environments, investigating adaptation dynamics when a self-interested adversary is introduced.  Finally, Scmid et al. \cite{Schmid2021LearningTP} explore using agents that can explicitly impose penalties in a zero-sum setting, applied in N-player Prisoner's Dilemma games with large agent populations.

Given this body of work, the contributions of this work are as follows:
\begin{itemize}
\item A betrayal-oriented environment:
  we design a simple, limited ruleset that can result in the emergence complex betrayal behaviors, consolidated in a single-agent RL environment.
\item Interpretable Betrayal Detection: Proposal of a classification-based detector that utilizes explainable, behavioral / observational evidence generated during agent play.
\item Betrayal penalization: proposal of avenues for penalizing detected betrayal during learning.
\item Experimental validation: we provide preliminary empirical findings showcasing emergence and successful detection of betrayal behaviors in the proposed environment.
\item Future work proposals: we suggest pathways for utilizing the rich potential of the environment in future work, ruleset extensions and
  additional investigation axes of interest.

\end{itemize}

\section{Proposed Environment}

The proposed environment is built with a focus on betrayal detection and penalization goals expressed in the literature \cite{armstrong2021ai}, extending previous work on agent communication in RL settings \cite{kajic2020learning}.

It implements an episodic game that consists of a collection of $N \geq 2$ gridworlds $[G_1\dots G_n]$, each paired with a single agent $A_i$. All worlds are associated with a pool of $k \geq N$ food items $F = [f_1, \dots, f_k]$ that provide variable reward and nutrition to agents upon consumption.
The environment advances in a single-agent, turn-based fashion, using the following rules and mechanics:

\begin{itemize}
\item The game is played in rounds, wherein all agents act once in a randomly generated order.
\item At the start of each round, food items are randomly allocated and positioned in each world.
\item The objective of each agent $A_i$ is to obtain food, which yields reward.
  Agent $A_i$ may harvest food by probing a location within their world $G_i$, but other worlds are inaccessible.
\item Agents cannot observe their own gridworld. Instead, they may observe all other (``opponent'') worlds and communicate with their respective opponent agents, conveying information of food locations within them -- i.e, agent $A_i$ sends $N-1$ messages $m_{ij} | j\in [1, \dots N], j \neq i$ with $m_{ij}  \in \mathbb{R}^d$ carrying information about where food is located in gridworld $G_j$, according to agent $A_i$.
\item Agents utilize incoming messages from other agents to decide where to probe / navigate for food within their world. If a food item is discovered in the destination and consumed, the agent obtains the reward amount it contains.
\item If an agent fails to consume food in a turn, they gain hunger. Hunger affects an agent's communication capabilities, distorting outgoing messages by a magnitude proportional to its value. The final transmitted message is $\hat{m}_{ij} = H(m_{ij})$, where $H(\cdot)$ is a noise function.
\item At the end of each round (i.e., once all agents have acted), if the food pool is empty, the episode ends. Otherwise, the procedure restarts with a new round.
\end{itemize}

This setting defines an social contract, where well-meaning agents are expected to truthfully relay food item coordinates for mutual benefit. However, interesting cases of betrayal also arise.
Namely, a deceptive agent $A_d$ may choose to transmit dishonest location coordinates: any food item in opponent worlds has a chance to be randomly relocated within $G_d$ in future rounds. At the same time, $A_d$ has to selectively regularize, cycle and/or distribute deception among their adversaries to avoid resorting to blind navigation: any systematically starved opponent agent will become unreliable in providing directions. Example illustrations of game mechanics and cases of betrayal / hunger trade-offs are available in the appendix, in figures \ref{fig:betrayal_hunger} and \ref{fig:overview} respectively.

\section{Preliminary Experiments}

\subsection{Betrayal Emergence}
In order to empirically test the potential of the proposed environment to produce cases of betrayal behaviors, we perform a set of preliminary experiments.
We train a configuration with $N=2$ girdworlds: the first agent (Alice) is trained from scratch, using the popular Proximal Policy Optimization (PPO) \cite{schulman2017proximal} algorithm with an MLP policy model. The learning agent is trained against an opponent (Bob) fixed to truthful behavior.
Food nutrition and reward are equalized for simplicity and sampled from distinct values in $[0, 1]$.
Bob is set to transmit one-hot location encodings, with non-zero values scaled proportionally to the observed food item rewards.
This configuration biases Alice towards adopting an interpretable messaging protocol, i.e. the same encoding scheme that Bob transmits and expects.
We apply hunger distortion of outgoing messages via additive uniform noise sampled from $[-h_i, h_i]$, where $h_i$ is the hunger value of the agent.

To determine whether an action from the $i$-th agent constitutes betrayal, we compare their intended message $m_{ij}$ (prior to any hunger-induced degradation) with true food locations in the observed world $G_j$. For $l = argmax(m_{ij})$, betrayal occurs when no food can be found in the $l$-th grid position ($G_j(l)=0$). 
Regarding implementation, we used python, gym \cite{gym} and stable-baselines3 \cite{sb3}. The developed environment will be made publicly available shortly, upon reaching a polished version.


\begin{figure}
  \centering
  \includegraphics[scale=0.25]{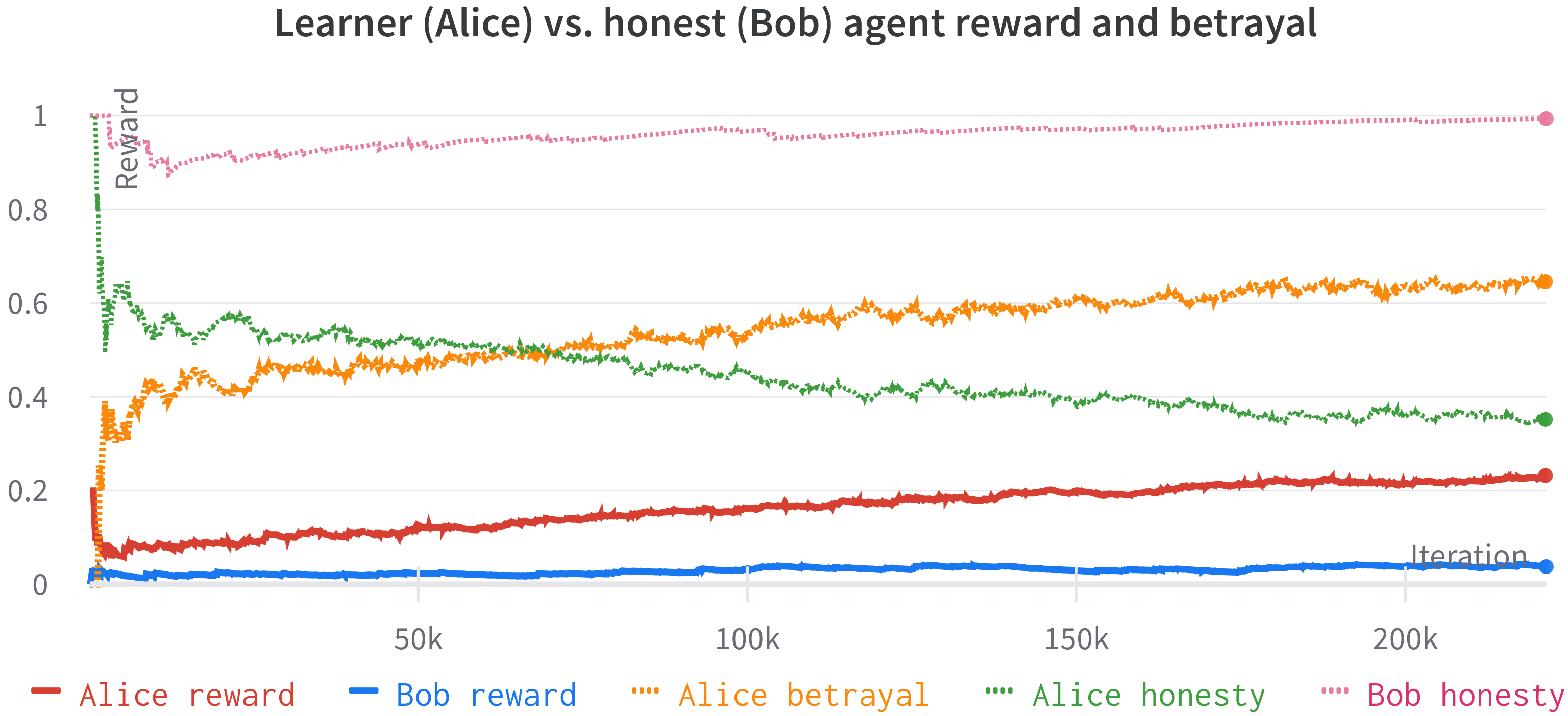}
  \caption[Betrayal Hunger]{Comparison between learner and honest agent rewards (red, blue), along with exhibited behavior of betrayal (orange) and honesty (dotted green, pink) for Alice and Bob respectively. Bob's betrayal score is 0 and is omitted. Values are illustrated with exponential moving average smoothing.}
  \label{fig:training}
\end{figure}

Figure \ref{fig:training} illustrates experimental results after training for $1e5$ timesteps, with a hunger increase delta of $0.15$ and a gridworld size of $5$ tiles. ``Honesty'' scores denote cases where incoming messages match true food locations (i.e. neither indented nor hunger-induced misdirection), and are progressively dominated by betrayal, given hunger and hunger-induced distortion results (see appendix Figure \ref{fig:hungers}).
Alice outperforms the honest baseline in terms of gained reward per step, with respective betrayal and honesty measurements rising and dropping as training progresses, respectively. In other words, Alice learns to adopt instrumentally useful actions of betrayal to obtain increased reward.
Bob's reward increases at a slower rate as Alice is learning, while betrayal is zero and honesty scores stay high, affected only by hunger (presumably reflecting Alice's increasing grasp on the communication protocol). We believe that these findings provide evidence for the potential of the environment for generating betrayal patterns; subsequent empirical work of larger scale and additional investigation axes should result in further useful results.

\subsection{Betrayal Detection and Penalization}

In order to facilitate betrayal detection and penalization, we apply a simple but intuitive approach.
We run the trained agent for $500$ episodes to collect interpretable run metadata, e.g. current / cumulative values for hungers, rewards, sent messages, etc. We use the resulting $4982 \times 33$ feature matrix and generated ground truth betrayal values
to train a feedforward neural network to predict the betrayal label. After hyperparameter tuning and $3$-fold cross validation,
we obtain a macro F1 mean scores of $68.35 \% $ (stdev $1.18 \%$),
compared to a probabilistic-based baseline of $49.36 \%$. This illustrates that betrayal detection in the proposed setting is possible, using explainable, interpretable features.

For disincentivization of deceptive behaviors, we are investigating utilizing classifier probability outputs as betrayal penalty modifiers applied during training in the agent's reward (current) or the policy learning loss (future).
This has proven challenging, as agents appear to game the classifier to near-zero betrayal scores, suggesting that the utilized feature set potentially fails to encapsulate deceptive behavior precisely. To this end, we are in the process of augmenting our penalization investigation to extraction of interpretable features sequences (e.g. metadata recent history), increased dataset sizes, and scaling up classifier models (e.g. attention-based networks). Additionally, we will employ feature selection to discard irrelevant features in an effort to remove degrees of freedom that the agent may use to hack / game penalization penalties during training.

\section{Conclusions and Future Work}

In this work we presented an RL environment
leveraging a communication-based cooperative navigation task,
geared towards detecting and penalizing betrayal.
Preliminary experimental results provide evidence for successful betrayal emergence and detection, while multiple avenues for classification-based betrayal penalization with interpretable features are proposed and currently pursued.

We believe that the proposed environment presents rich potential for generating interesting social interactions patterns, which could be valuable research topics in future studies.
Such work includes exploration of different betrayal dynamics (e.g. utilizing passive reward penalties, learning versus a dishonest opponent, opponents with different capability / penalization attributes) and tracking of onset and evolution of betrayal patterns like reciprocity, defection and retribution.
Higher-level patterns may include tacticaly play and long-term strategizing, e.g. prioritizing reward-rich food for self consumption while reserving high-nutrition food for adversaries, limiting opponent hunger under a certain threshold, ramping up betrayal when food becomes scarce, etc.
Other avenues include measuring the effect of different environmental properties and axes (e.g. world dimensionality, food abundance, food distribution during relocation, etc.) to observed dynamics.

Moreover, interesting extensions to this work are examining emergent behaviors under more sophisticated opponent modeling (e.g. as in related work \cite{kim2020communication}) or focusing on different dishonesty patterns (e.g. ``accidental'' deception derived from hunger or under/overfitted opponent).
Finally, a natural extension of the proposed work involves exploring the effect of consensus and trustworthiness on betrayal dynamics when dealing with multiple rather than a single adversary (e.g. cycling betrayal victims, prioritizing deception to unreliable communicators, etc.), or adopting a multiagent approach to the environment for simultaneous rather than turn-based play.

\begin{ack}
This work has been supported by the Effective Altruism Long-Term Future Fund \footnote{\url{https://funds.effectivealtruism.org/funds/far-future}}. Additionally, valuable feedback was provided by Tim Farrelly and Quntin Pope via the AI Safety Camp program \footnote{\url{https://aisafety.camp}}, as well as Stuart Armstrong, who also proposed the project this work is based on \footnote{\url{https://www.alignmentforum.org/s/xujLGRKFLKsPCTimd/p/oeCXS2ZCn4rPyq7LQ}}.
\end{ack}

\bibliographystyle{acm}
\bibliography{references.bib}

\newpage
\appendix

\section*{Appendix: Figures}

\begin{figure}[h]
  \centering
 \includegraphics[scale=0.4]{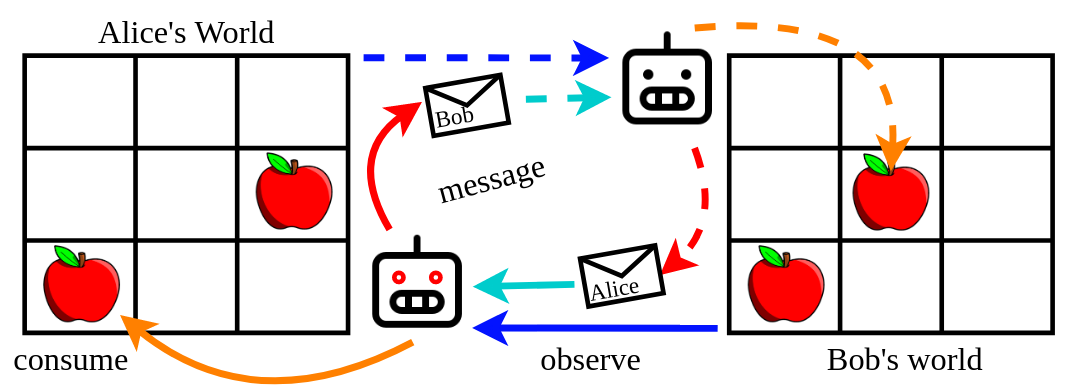}
  \caption[Overview]{An overview of the proposed environment for investigating betrayal for the example case of two agents, Alice (left, solid lines) and Bob (right, dashed lines). An agent's turn consists of a composite observation space of opponent worlds (blue) and incoming messages (cyan). The action space involves composing a message to other agents (red) and navigating the owned gridworld to obtain food (orange).}
  \label{fig:overview}
\end{figure}

\begin{figure}[h]
  \centering
  \includegraphics[scale=0.45]{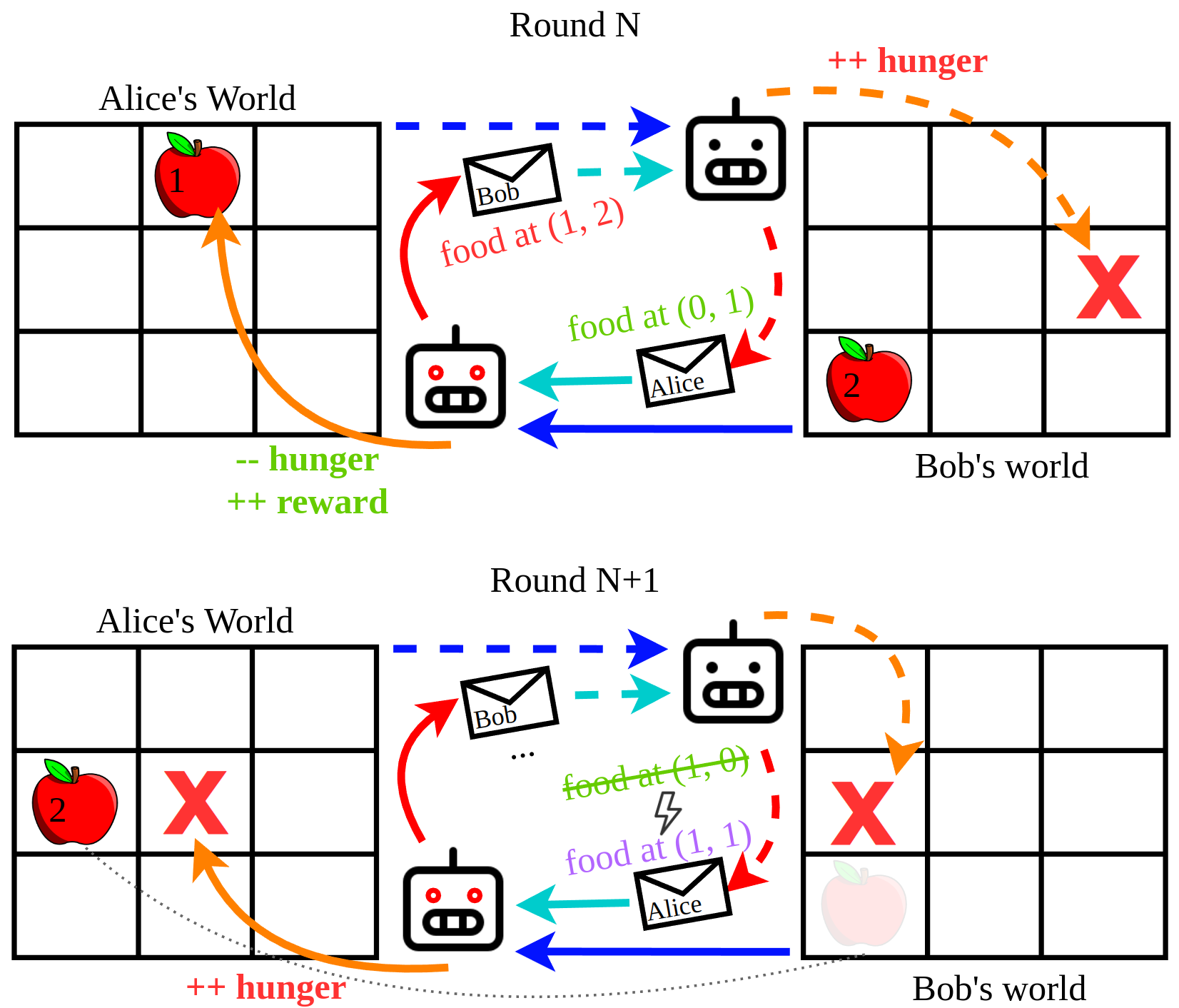} 
  \caption[Betrayal Hunger]{Top: Bob engages in honest communication, enabling Alice to consume food \#1, obtain reward and lose hunger. Alice betrays by transmiting false coordinates to food item \#2 in Bob's world, which remains intact and survives to the next round. Botttom: In the next round, the preserved food item \#2 randomly relocates to Alice's world, enabling her to capitalize on their betrayal. However, Bob has been betrayed too many times to reliably communicate, resulting in their message being corrupted and Alice suffering hunger penalties in turn.}
  \label{fig:betrayal_hunger}
\end{figure}

\begin{figure}[h]
  \centering
  \includegraphics[scale=0.25]{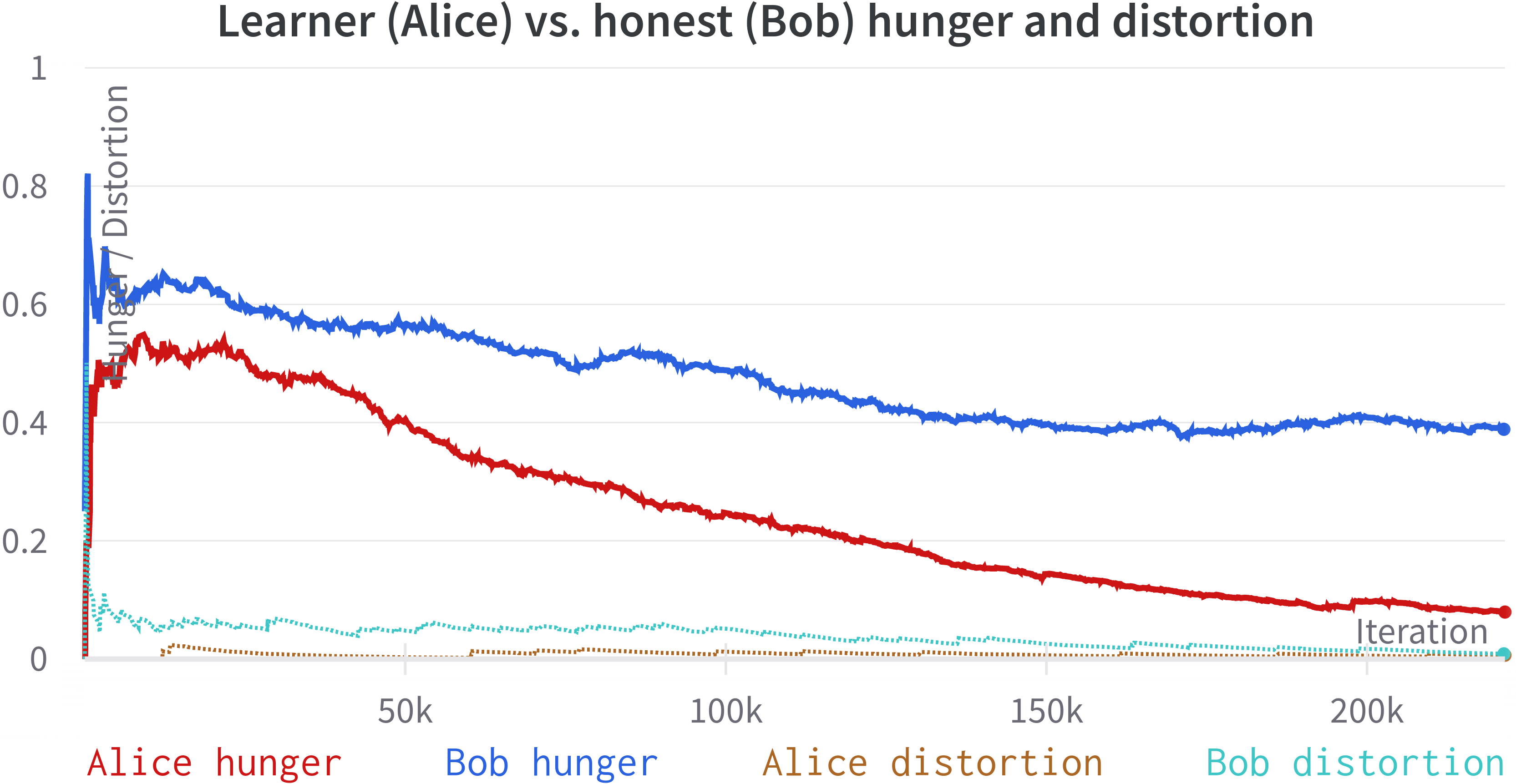} 
  \caption[Betrayal Hunger]{Hunger scores (red, blue) and hunger-induced message distortion (brown, cyan) per time step during training, for Alice and Bob respectively. Decreasing scores for hunger and hunger-based distortion indicate that opponents improve in transmitting messages with true food locations, for both agents.}
  \label{fig:hungers}
\end{figure}

\end{document}